\documentclass{article}
\usepackage{ijcai22}

\usepackage{times}

\usepackage{soul}
\usepackage{url}
\usepackage{pdfpages}
\usepackage[hidelinks]{hyperref}
\usepackage[utf8]{inputenc}
\usepackage[small]{caption}
\usepackage{graphicx}
\usepackage{amsmath}
\usepackage{booktabs}
\usepackage{multirow}
\usepackage{xcolor, colortbl}
\usepackage{algorithm}  
\usepackage{algorithmicx} 
\usepackage{algpseudocode}

\newcommand{\mc}[1]{\mathcal{#1}}

\definecolor{Gray}{gray}{0.9}
\newcolumntype{a}{>{\columncolor{Gray}}c}
\newcommand*\rot{\rotatebox{90}}

\def\eg{{\em e.g.}}
\def\ie{{\em i.e.}}

\newtheorem{defn}{Definition}

\title{Enhancing the Transferability via Feature-Momentum Adversarial Attack}

\author{
Xianglong He\and
Yuezun Li\footnote{Corresponding author}\and
Haipeng Qu\And
Junyu Dong\\
\affiliations
Ocean University of China\\
}

\begin{document}

\maketitle

\begin{abstract}
Transferable adversarial attack has drawn increasing attention due to their practical threaten to real-world applications. In particular, the feature-level adversarial attack is one recent branch that can enhance the transferability via disturbing the intermediate features. The existing methods usually create a guidance map for features, where the value indicates the importance of the corresponding feature element and then employs an iterative algorithm to disrupt the features accordingly. However, the guidance map is fixed in existing methods, which can not consistently reflect the behavior of networks as the image is changed during iteration. In this paper, we describe a new method called Feature-Momentum Adversarial Attack (FMAA) to further improve transferability. The key idea of our method is that we estimate a guidance map dynamically at each iteration using momentum to effectively disturb the category-relevant features. Extensive experiments demonstrate that our method significantly outperforms other state-of-the-art methods by a large margin on different target models. 
\end{abstract}

\section{Introduction}

Deep Neural Networks (DNNs) have shown impressive performance in a variety of computer vision tasks, such as image classification \cite{simonyan2014very}, autonomous driving \cite{zeng2020dsdnet,janai2020computer}, scene parsing \cite{long2015fully,he2017mask} and etc \cite{wang2020deep,zhang2016single}. However, it has been proven that DNNs are vulnerable to adversarial perturbations \cite{szegedy2014intriguing}, which are intentionally crafted and imperceptible noise, yet can disrupt the model predictions. Adversarial attacks have drawn increasing attentions in recent years, as it raises severe concern in modern DNN-based models. Meanwhile, studying adversarial attacks can also help improve the model robustness \cite{goodfellow2015explaining,madry2017towards}.   

Adversarial attacks can be divided into two categories of white-box attack and black-box attack, according to how much knowledge is revealed to attackers. In the white-box setting, the attackers can access the details of target models, \eg, the architecture and weight parameters \cite{szegedy2014intriguing,madry2017towards}. Thus the attackers can optimize a designed objective with respect to image and create adversarial perturbations overfitted to the target model. In contrast, the black-box attack does not require the details of the target model. One typical solution for the black-box attack is called transferable adversarial attack, which focuses on improving the transferability of adversarial perturbations, \ie, the adversarial perturbations created on one model can attack other different models. \cite{papernot2016transferability,yang2020learning}. Compared to the white-box attack, the transferable adversarial attack is more practical in real-world applications. 

\begin{figure}[t]
\centering
\includegraphics[width=\linewidth]{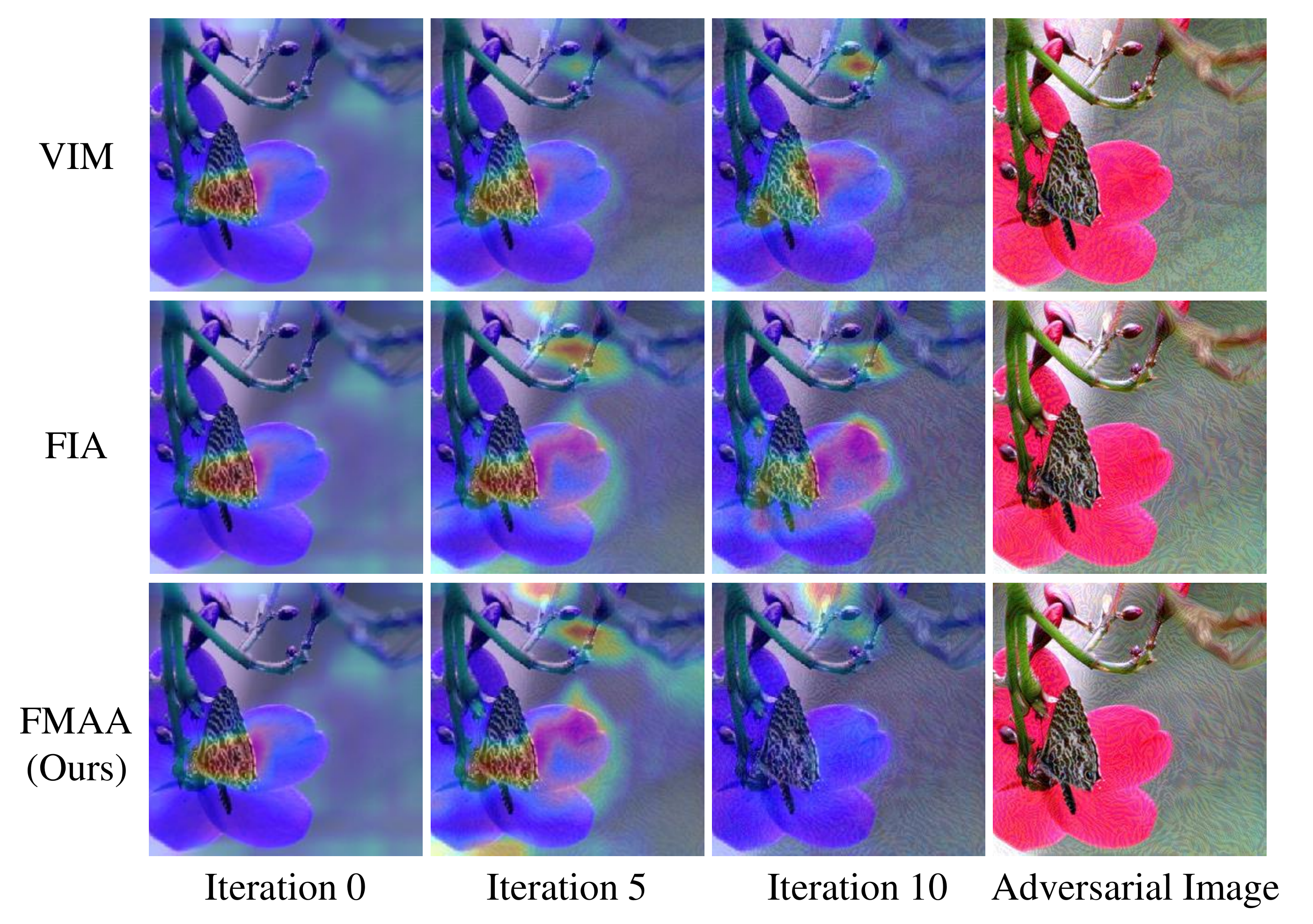}
\vspace{-0.7cm}
\caption{\small Visual comparison of our method (bottom row) and most recent two state-of-the-art methods VIM \protect \cite{wang2021enhancing} and FIA \protect \cite{wang2021feature} (top two rows). The source model is Inc-v3 and target model is Vgg-16. The left three columns show the Grad-CAM \protect \cite{selvaraju2017grad} of features attacked by these methods and the rightmost column is corresponding adversarial images. We can see that our method can significantly disturb the category-relevant features. More examples can be found in Supplementary.}
\label{fig:teaser}
\vspace{-0.3cm}
\end{figure}

To date, many works are proposed to improve the transferability of adversarial perturbations by meticulous adaptation of gradient \cite{dong2018boosting,lin2019nesterov,wang2021enhancing}, or employing random input transformations \cite{xie2019improving,dong2019evading}. Recently, a vein of works \cite{zhou2018transferable,huang2019enhancing,ganeshan2019fda,wang2021feature} directly disturbs the intermediate features instead of the model output, so as to mitigate the overfitting effect introduced by task-specific objectives. In general, these methods utilize a guidance map for feature disturbance, where each value corresponds the disturbing strength of corresponding feature element. Since the feature is complex, how to effectively attack features is still an open question.

In this work, we propose a new feature-level method called Feature-Momentum Adversarial Attack (FMAA), which is an iterative approach that attacks the features under dynamic guidance based on momentum. Note that the feature contains both category-relevant features (\eg, objects) and category-irrelevant features (\eg, backgrounds) \cite{he2016deep}, and the category-relevant feature is highly generalized across models compared to the category-irrelevant feature \cite{wang2021feature}. To effectively disturb the category-relevant feature, our method generates a dynamic guidance map instead of a fixed guidance map used in existing methods. The motivation is that the image is changed after each iteration, resulting in an alteration of features consequently. The key idea of our method is that we adapt the momentum formulation into the generation of guidance map, \ie, the current guidance map considers the knowledge from previous iterations. In this way, the guidance map is more comprehensive and can describe the category-relevant feature more precisely, leading to a significant improvement in transferability (See Fig. \ref{fig:teaser}).

Extensive experiments are conducted on the ImageNet dataset \cite{deng2009imagenet} with the comparison against seven state-of-the-art attack methods on nine original models and five defense models. The experiments corroborate the superiority of our method, which notably outperforms others on attacking original and defense models respectively. Moreover, we conduct ablation studies on the effect of different parameters regarding the feature momentum, aiming to provide more insight for the following research.

\section{Related Works}

Adversarial attack can be categorized into two classes, white-box attack and black-box attack. Transferable adversarial attack is one recent typical branch to solve black-box attack and our method falls into this class. We briefly introduce each class of adversarial attack in the following.


\smallskip
\noindent{\bf White-box Adversarial Attack.}
In this setting, attackers can access the model properties such as parameters and architectures. Thus the attackers can craft an objective and optimize it using the gradient back-propagation to fool the prediction of models, \eg, \cite{szegedy2014intriguing,goodfellow2015explaining,kurakin2016adversarial,carlini2017towards}. These methods formulate adversarial attack as an optimization problem under constraints on the distortion of perturbations. They either use one-step or iterative steps fashion to alter the image. Since the objective is crafted on a known model, the attack is overfitted on this model, which thereby performs badly on other models. The white-box adversarial attack is not practical for real-world applications as the attackers usually can not know the details of models. 
 
\smallskip
\noindent{\bf Transferable Adversarial Attack.}
In contrast to white-box attack, transferable adversarial attack aims to improve the transferability of adversarial attack such that it can attack different models without knowing the details. To date, many methods are proposed to improve transferability.  Several works mitigate the overfitting phenomenon using advanced gradients, such as utilizing momentum to update the gradients in each iteration \cite{dong2018boosting}, proposing Nesterov accelerated gradient \cite{lin2019nesterov} and gradient variance \cite{wang2021enhancing} to boost transferability. Other works such as \cite{xie2019improving,dong2019evading} increasing the diversity of input images to improve transferability, where the work of \cite{xie2019improving} employed random image translation and the work of \cite{dong2019evading} optimizing objectives on translated images.  

Note the aforementioned works focus on disturbing the output of the model. There is another vein of works are dedicated to attacking the intermediate features. The work of \cite{zhou2018transferable} directly enlarged the distance between attacked features and their original ones. The work of \cite{huang2019enhancing} improved the transferability by increasing its perturbation on a pre-selected layer of the source model. The work of \cite{ganeshan2019fda} disrupted the feature at each layer of the network to achieve high transferability. More recently, the work of \cite{wang2021feature} proposed feature importance-aware attack, which provides a guidance map to indicate important areas of the feature. The works \cite{dong2018boosting,wang2021feature} are most relevant to us as the first work utilizes the momentum to the optimization of the objective and the second one disturbs the feature with specific guidance. The key difference to the first work is the feature momentum we proposed is used to improve the accuracy of the guidance map for feature disturbance, which can be cooperated with the momentum used in the optimization of the objective. Compared to the work of \cite{wang2021feature} that fixes the guidance map before attack, our work proposes a dynamic guidance map, which absorbs more knowledge in previous iterations and thereby is more effective than other counterparts.   

\section{Methodology}
In this section, we describe the details of the proposed Feature-Momentum Adversarial Attack (FMAA). Our method is motivated by that the category-relevant feature is more general across models and considering the knowledge over iterations can better guide the disturbance of category-relevant features.  

\subsection{Problem Formulation}
Denote $\mc{F}: x \rightarrow y$ as the mapping function of classifier $\mc{F}$, where $x$ denotes the clean input image, $y$ denotes its corresponding true label and $\theta$ is the parameters of the classifier. Our goal is to seek the adversarial image $x^{adv}$, which has a small distortion compared to the clean image $x$, but can make the prediction go wrong, \ie, $\mc{F}(x^{adv}; \theta) \neq y$. Denote $\mc{L}$ as the original objective function used in training of classifier (\eg, cross-entropy). Thus the problem of generating adversarial image $x_{adv}$ can be formulated as 
\begin{equation}
    \mathop{\arg\max}_{x^{adv}} \ \ \mc{L}(x^{adv}, y; \theta), \ \ \textrm{s.t. } ||x^{adv} - x||_{\infty} \leq \epsilon,
\end{equation}
where $\epsilon$ is the budget of adversarial perturbations. Note in the black-box setting, the parameters $\theta$ are unknown to attackers, which obstacles the optimization of the above equation. The transferable adversarial attack can avoid this problem by disturbing a source model that the parameters can be accessed by attackers, and then improving the transferability of adversarial perturbations to attack the target model $\mc{F}$. Typically, the generation of transferable adversarial perturbation uses iterative steps to optimize the above equation (\eg, MIM \cite{dong2018boosting}, FIA \cite{wang2021feature}, etc).

\subsection{Feature-Momentum Adversarial Attack}
Our method focuses on generating a dynamical guidance map that leads to disturbing the category-relevant features. Intuitively, we can utilize the gradient with respect to the feature as the guidance map to indicate category-relevant features. 
Suppose that we attack the $i$-th layer of classifier $\mc{F}$ and denote the feature of $i$-th layer as $\mc{F}_{i}(x; \theta)$. Thus the gradient on this feature can be denoted as $\nabla_{\mc{F}_{i}(x; \theta)} \mc{L}(x, y; \theta)$.
Similar to existing methods, our method also follows the iterative way to generate adversarial perturbations.
In this scenario, a fixed guidance map can not fully represent the category-relevant feature as the image is changed after each iteration. Therefore, we propose feature momentum to generate a more precise guidance map at each iteration.

\begin{defn}
{\bf \em Feature Momentum.} Given an input image $x$, a classifier $\mc{F}$ with parameters $\theta$ and the objective function $\mc{L}$, the feature momentum of $i$-th layer at $t$-th iteration can be defined as 
\begin{equation}
    \beta_{t+1} = \lambda \cdot \beta_{t} + \frac{\nabla_{\mc{F}_{i}(x_t; \theta)} \mc{L}(x_t, y; \theta)}{||\nabla_{\mc{F}_{i}(x_t; \theta)} \mc{L}(x_t, y; \theta)||_{1}}.
\end{equation}
\end{defn}
In this way, the guidance map is updated partially based on the knowledge at previous iterations, thus it can represent the category-relevant feature consistently with iteration going. Since the gradients calculated on one specific model can contain the traces specific to the model itself, we employ an aggregation strategy \cite{wang2021feature} to reduce those traces. Specifically, we utilize random transformation (\eg, masking) on the input image to create several variants and then average the gradients over these variants as the final gradients. Thus the gradients at $i$-th layer can be rewritten as $\frac{1}{n} \sum \nabla_{\mc{F}_{i}(\mc{T}(x); \theta)} \mc{L}(\mc{T}(x), y; \theta)$, where $\mc{T}(\cdot)$ denotes the random transformation applied on images.

\begin{figure}[t]
\centering
\includegraphics[width=\linewidth]{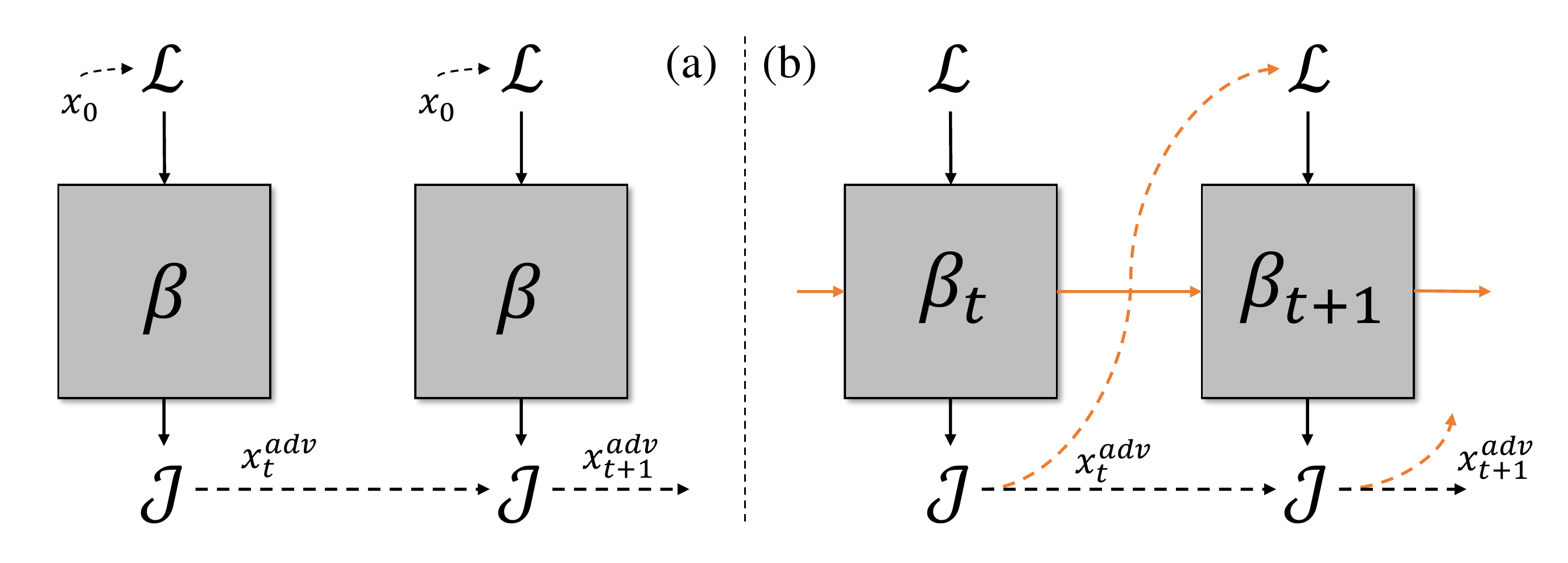}
\vspace{-0.7cm}
\caption{\small The comparison of (a) existing methods and (b) our method. $\beta$ denotes guidance map. Our method dynamically updates the guidance map using feature momentum.}
\label{fig:method}
\vspace{-0.3cm}
\end{figure}

Given the guidance map, we can design objective functions to disturb features more effectively. The higher value in the guidance map indicates the spot in the feature is more critical. Thus spending more effort to disturb these spots can greatly affect the feature. Then the objective function can be defined as  
\begin{equation}
    \mc{J}(x^{adv}; x, \theta, \beta) = \sum \beta \odot \mc{F}_{i}(x^{adv}; \theta),
\end{equation}
where $\odot$ denotes the element-wise multiplication. We utilize MIM \cite{dong2018boosting} to optimize this objective. The algorithm procedure is shown in Alg. \ref{alg}. The comparison of our method and existing methods is shown in Fig. \ref{fig:method}.

\begin{algorithm}[ht!]
\caption{\small Feature Momentum Adversarial Attack}
\label{alg}
\begin{algorithmic}
\Require classifier $\mc{F}$ with parameters $\theta$; input image $x$; perturbation budget $\epsilon$; maximum iteration number $T$; feature-momentum weight $\lambda$.
\Ensure adversarial image $x^{adv}$.
\vspace{0.1cm}
\State $t=0, \beta_0 = 0, g_0 = 0, x^{adv}_0 = x$
\While{$t<T$}
\State // Updating feature momentum for feature disturbance
\State $\beta_{t+1} = \lambda \cdot \beta_{t} + \frac{\nabla_{\mc{F}_{i}(x^{adv}_t; \theta)} \mc{L}(x^{adv}_t, y; \theta)}{||\nabla_{\mc{F}_{i}(x^{adv}_t; \theta)} \mc{L}(x^{adv}_t, y; \theta)||_{1}}$
\State // Updating gradient momentum for objective optimization
\State $g_{t+1} = \mu \cdot g_{t} + \frac{\nabla_{x^{adv}_t} \mc{J}(x^{adv}_t; x, \theta, \beta_{t+1})}{||\nabla_{x^{adv}_t} \mc{J}(x^{adv}_t; x, \theta, \beta_{t+1})||_{1}}$
\State $x^{adv}_{t+1} = \text{Clip}_{\epsilon}\{x^{adv}_{t} - \alpha \cdot \text{sign}(g_{t+1}) \}$
\State $t=t+1$
\EndWhile
\end{algorithmic}
\end{algorithm}


\section{Experiments}

\subsection{Experimental Settings}

\smallskip
\noindent{\bf Dataset.} We evaluate the proposed method on the ImageNet-compatible dataset \cite{deng2009imagenet,kurakin2018adversarial} as in previous works \cite{wang2021feature}. This dataset is used in NIPS 2017 adversarial competition, including $1000$ images in total.


\smallskip
\noindent{\bf Compared Attack Methods.} Our method is compared with various state-of-the-art attacks, \eg,  MIM \cite{dong2018boosting}, DIM \cite{xie2019improving}, TIM\cite{dong2019evading}, PIM \cite{gao2020patch}, NRDM \cite{naseer2018task}, FDA \cite{ganeshan2019fda}, FIA \cite{wang2021feature} and VIM \cite{wang2021enhancing}. Moreover, we also compare our method with their variants, \eg, the combination of TIM and DIM as TIDIM, the combination of PIM and DIM as PIDIM, and the combination of PIM, TIM and DIM as PITIDIM. The settings of each method are described in Supplementary.

\smallskip
\noindent{\bf Models.} Our method is validated on several models, covering different original and defense models. The original models contain Inception-V3 (Inc-v3) \cite{szegedy2016rethinking}, Inception-V4 (Inc-v4) \cite{szegedy2017inception}, Inception-ResNet-V2 (IncRes-v2), ResNet-V1-50 (Res-v1-50) \cite{he2016deep}, ResNet-V1-152 (Res-v1-152), ResNet-V2-50 (Res-v2-50), ResNet-V2-152 (Res-v2-152), Vgg-16  \cite{simonyan2014very} and Vgg-19. The defense models are adversarially trained \cite{kurakin2016adversarial,tramer2017ensemble}, which includes Adv-Inc-v3, Adv-IncRes-v2, Ens3-Inc-v3, Ens4-Inc-v3, and Ens-IncRes-v2. For all the experiments, we select Inc-v3, IncRes-v2, Res-v1-152, vgg-16 as source models and attack all of the models. 

\smallskip
\noindent{\bf Implementation Details.}
The setting of parameters are as follows: The perturbation budget $\epsilon=16$, the maximum iteration number $T=10$, the step size $\alpha = \epsilon/T$, the feature-momentum weight $\lambda=1.1$, the general momentum weight $\mu = 1$. For random transformation of $\mc{T}$, we use random pixel masking as in FIA \cite{wang2021feature}, where each pixel has a dropping probability of $p$. In our case, we set the dropping probability of start iteration as $0.4$ and use $0.1$ for the rest of the iterations. The number of transformation is set to $n = 30$. The layer chosen for the attack is set as follows: we select Conv\_4a for Inc-V3, Conv3\_3 for Vgg-16, Conv\_4a for IncRes-V2, and the last layer of the second block for Res-v1-152. 

\renewcommand\arraystretch{0.9}
\begin{table*}[!ht]
    \centering
    \small
    \caption{\small Success rate ($\%$) of different methods attacking different original models. Source models are listed in the first column and target models are listed in the first row.}
    \vspace{-0.3cm}
    \resizebox{\linewidth}{!}{
    \begin{tabular}{c|c|c|c|c|c|c|c|c|c|c}
    \toprule
    ~ & Attack & Inc-v3 & Inc-v4 & IncRes-v2 & Res-v1-50 & Res-v1-152 & Res-v2-50 & Res-v2-152 & Vgg-16 & Vgg-19  \\ 
    \midrule    
    \multirow{11}{*}{Inc-v3} & MIM & \textbf{100.0} & 42.5 & 38.8 & 39.0 & 32.8 & 34.7 & 32.4 & 40.0 & 38.3 \\ 
    ~ & DIM & 99.0 & 64.5 & 61.9 & 53.1 & 47.5 & 54.8 & 51.8 & 53.0 & 52.8 \\ 
    ~ & PIM & 100.0 & 49.2 & 47.3 & 46.4 & 38.5 & 42.8 & 37.1 & 51.3 & 48.7 \\ 
    ~ & NRDM & 81.3 & 52.6 & 49.8 & 51.4 & 39.7 & 56.8 & 43.7 & 44.8 & 44.6 \\ 
    ~ & FDA & 74.4 & 46.8 & 41.4 & 38.9 & 32.2 & 50.1 & 38.5 & 33.3 & 33.7 \\ 
    ~ & FIA & 97.8 & 80.2 & 78.2 & 73.9 & 67.5 & 77.1 & 71.4 & 73.3 & 74.3 \\ 
    ~ & VIM & 100.0 & 70.3 & 66.9 & 57.7 & 51.5 & 58.1 & 55.6 & 57.9 & 53.1 \\
    ~ & \bf FMAA & 99.6 & \bf 90.7 & \bf 89.6 & \bf 83.2 & \bf 77.9 & \bf 86.7 & \bf 83.9 & \bf 81.9 & \bf 82.2 \\ 
    \cmidrule{2-11}  
    ~ & PIDIM & 99.9 & 71.9 & 67.5 & 58.5 & 52.2 & 60.4 & 53.0 & 63.0 & 61.0 \\ 
    ~ & FIA+PIDIM & 96.9 & 82.7 & 81.0 & 81.7 & 77.6 & 80.8 & 75.7 & 80.6 & 81.9 \\
    ~ & \bf FMAA+PIM & \bf 99.8 & \textbf{93.5} & \textbf{92.6} & 87.6 & 83.6 & 88.4 & \textbf{87.3} & 86.4 & 86.6 \\ 
    ~ & \bf FMAA+PIDIM & 99.5 & 92.2 & 92.2 & \textbf{90.2} & \textbf{87.1} & \textbf{89.3} & 86.6 & \textbf{86.6} & \textbf{88.2} \\ 
    \midrule    
    \multirow{11}{*}{IncRes-v2} & MIM & 60.4 & 51.1 & 99.0 & 46.2 & 39.7 & 48.9 & 42.8 & 47.3 & 42.5 \\ 
    ~ & DIM & 73.0 & 66.0 & 95.4 & 58.0 & 53.6 & 60.6 & 58.6 & 56.5 & 53.3 \\ 
    ~ & PIM & 65.2 & 58.4 & \bf 99.3 & 51.7 & 45.7 & 52.7 & 46.7 & 57.1 & 55.5 \\
    ~ & NRDM & 70.6 & 67.4 & 77.5 & 64.6 & 52.7 & 73.0 & 59.7 & 53.3 & 52.4 \\ 
    ~ & FDA & 69.2 & 68.0 & 78.2 & 62.1 & 50.1 & 73.8 & 56.2 & 49.2 & 44.8 \\ 
    ~ & FIA & 80.1 & 77.7 & 89.1 & 73.7 & 68.5 & 75.3 & 71.1 & 73.1 & 73.0 \\ 
    ~ & VIM & 78.7 & 76.0 & 99.0 & 66.2 & 61.5 & 70.0 & 66.3 & 64.2 & 61.3 \\
    ~ & \bf FMAA & \bf 90.5 & \bf 87.5 & 97.0 & \bf 83.0 & \bf 80.3 & \bf 85.6 & \bf 83.9 & \bf 82.1 & \bf 81.8 \\ 
    \cmidrule{2-11} 
    ~ & PIDIM & 80.6 & 77.6 & 97.9 & 65.2 & 60.5 & 69.2 & 65.4 & 67.2 & 66.6 \\ 
    ~ & FIA+PIDIM & 84.7 & 80.4 & 91.6 & 82.2 & 78.2 & 80.9 & 75.9 & 81.1 & 80.5 \\ 
    ~ & \bf FMAA+PIM & \textbf{94.3} & \textbf{93.1} & \textbf{99.4} & 89.4 & 87.0 & \textbf{90.6} & \textbf{88.9} & 87.5 & 87.2 \\ 
    ~ & \bf FMAA+PIDIM & 93.9 & 91.7 & 97.8 & \textbf{90.7} & \textbf{88.7} & 89.8 & 88.7 & \textbf{88.3} & \textbf{88.9} \\ 
    \midrule      
    \multirow{11}{*}{Res-v1-152} & MIM & 59.4 & 52.6 & 52.3 & 90.9 & 99.8 & 68.8 & 66.1 & 73.8 & 71.0 \\ 
    ~ & DIM & 79.8 & 69.1 & 70.1 & 93.5 & 99.8 & 81.8 & 80.7 & 84.0 & 84.2 \\ 
    ~ & PIM & 66.7 & 57.0 & 55.4 & 92.7 & 99.8 & 72.2 & 70.2 & 79.7 & 79.3 \\ 
    ~ & NRDM & 70.6 & 66.1 & 53.7 & 88.4 & 96.1 & 76.4 & 63.2 & 78.1 & 78.3 \\ 
    ~ & FDA & 70.6 & 66.8 & 56.2 & 88.7 & 94.2 & 78.7 & 67.2 & 76.7 & 77.8 \\ 
    ~ & FIA & 86.2 & 83.2 & 81.7 & 96.8 & 99.5 & 90.7 & 87.4 & 90.2 & 89.8 \\ 
    ~ & VIM & 79.9 & 72.7 & 73.2 & 94.9 & 99.9 & 85.8 & 84.4 & 86.4 & 84.2 \\
    ~ & \bf FMAA & \bf 92.6 & \bf 88.9 & \bf 88.7 & \bf 98.6 & \bf 99.8 & \bf 94.4 & \bf 92.2 & \bf 94.5 & \bf 93.7 \\ 
    \cmidrule{2-11} 
    ~ & PIDIM & 83.1 & 74.7 & 76.2 & 96.1 & 99.9 & 87.3 & 85.8 & 88.5 & 89.5 \\ 
    ~ & FIA+PIDIM & 92.8 & 88.5 & 87.9 & 97.7 & 99.5 & 93.6 & 91.5 & 95.2 & 94.5 \\     
    ~ & \bf FMAA+PIM & 95.4 & 91.9 & 91.0 & \textbf{99.5} & 99.8 & 96.4 & 95.4 & 97.4 & 96.9 \\ 
    ~ & \bf FMAA+PIDIM & \textbf{95.6} & \textbf{93.3} & \textbf{93.3} & 99.0 & \textbf{100.0} & \textbf{96.5} & \textbf{96.4} & \textbf{97.4} & \textbf{97.5} \\ 
    \midrule    
    \multirow{11}{*}{Vgg-16} & MIM & 82.7 & 82.4 & 75.9 & 88.6 & 83.8 & 83.6 & 78.6 & \textbf{100.0} & 97.0 \\ 
    ~ & DIM & 88.4 & 87.4 & 80.9 & 90.4 & 88.2 & 88.3 & 82.6 & 100.0 & 98.6 \\ 
    ~ & PIM & 87.0 & 86.6 & 82.3 & 90.7 & 87.4 & 88.3 & 82.6 & 100.0 & 97.4 \\ 
    ~ & NRDM & 79.7 & 77.5 & 61.7 & 78.8 & 73.8 & 74.4 & 66.6 & 92.7 & 91.7 \\ 
    ~ & FDA & 80.8 & 81.7 & 68.1 & 80.2 & 78.2 & 75.8 & 69.4 & 94.5 & 95.2 \\ 
    ~ & FIA & 96.2 & 96.7 & 93.0 & 96.8 & 94.0 & 95.4 & 92.3 & 99.9 & 99.5 \\ 
     ~ & VIM & 87.3 & 87.8 & 82.9 & 91.6 & 88.7 & 88.0 & 83.7 & 100.0 & 96.4 \\
    ~ & \bf FMAA & \bf 98.0 & \bf 98.4 & \bf 95.6 & \bf 98.4 & \bf 96.6 & \bf 97.1 & \bf 95.8 & 99.9 & \bf 99.5 \\ 
    \cmidrule{2-11} 
    ~ & PIDIM & 89.8 & 89.6 & 84.8 & 93.6 & 91.3 & 89.9 & 86.1 & 100.0 & 98.3 \\ 
    ~ & FIA+PIDIM & 97.1 & 97.5 & 95.1 & 97.7 & 95.7 & 96.4 & 94.3 & 100.0 & 99.6 \\ 
    ~ & \bf FMAA+PIM & 98.6 & \textbf{98.7} & 97.1 & 98.7 & 97.6 & 97.9 & 96.7 & 99.9 & \textbf{99.7} \\ 
    ~ & \bf FMAA+PIDIM & \textbf{98.7} & 98.4 & \textbf{97.3} & \textbf{99.2} & \textbf{97.7} & \textbf{98.2} & \textbf{97.1} & \textbf{100.0} & 99.6 \\ 
    \bottomrule
    \end{tabular}}
    \label{tab:attack_originals}
    \vspace{-0.4cm}
\end{table*}

\subsection{Results}
\smallskip
\noindent{\bf Attacking Original Models.}
Table \ref{tab:attack_originals} shows the attack performance of different methods on various original target models. The leftmost column denotes the source models and the top row denotes the target models. The same of source and target model represents white-box attack and different ones represent black-box attack.
From Table \ref{tab:attack_originals} we can observe that the proposed FMAA method outperforms all other attack methods by a large margin on black-box attack settings, and also achieves high attack success rates on white-box attack settings. Specifically, the attack performance of our method using Inc-v3, IncRes-v2, Res-v1-152 and Vgg-16 are $86.2\%, 85.7\%, 93.7\%, 97.7\%$ on average over all target models, which is improved by approximate $9.1\%, 10.0\%, 4.2\%, 1.7\%$ compared to the most recent method FIA. The significant improvement demonstrates the effectiveness of our method on transferability. We can also observe that the attack performance using Vgg-16 achieves the best performance compared to other more complex source models (over $95\%$), which is consistent with the analysis in work \cite{wang2021feature} that the complexity of the source model can affect the transferability.  

In addition, we also evaluate the performance of different combined methods. In particular, our method is combined with PIM and PIDIM in order and compared with the combined methods PIDIM, FIA+PIDIM. The results show that with the combination of PIM or PIDIM, our method can further boost the attack success rate on all target models and achieve the best performance compared to others. \ie, $90.5\%, 91.4\%, 96.6\%, 98.5\%$ on average, which outperforms FIA+PIDIM by approximate $8.4\%, 9.7\%, 3.1\%, 5.0\%$ respectively.

\smallskip
\noindent{\bf Attacking Defense Models.}
Table \ref{tab:attack_defense} shows the attack performance of different methods on various defense models. As these models are adversarially trained, they can resist the adversarial attacks, which result in a degradation of attack success rate compared to attacking original models. We can observe that our method still achieves the best attack performance than others, which are $53.3\%, 66.8\%, 79.4\%, 92.8\%$, improved by $6.7\%, 18.9\%, 4.4\%, 2.2\%$ on average compared to FIA. Note TIM is dedicated to attacking defense models, thus we do not include it in Table \ref{tab:attack_originals}. Similar to the above analysis, the attack performance using Vgg-16 is greatly higher than other source models, which represents the defense models likely obey the relationship between the complexity of models and transferability. 
For the comparison with combined methods, our method with PIM, PIDIM, PITIM and PITIDIM can still achieve the best performance. The results of this Table show a similar trend as in Table \ref{tab:attack_originals}, which corroborates the superiority of our method on transferability.


\renewcommand\arraystretch{0.9}
\begin{table}[!ht]
    \centering
    \small
    \caption{\small Success rate ($\%$) of different methods attacking different defense models. Source models are listed in the first column and target models are listed in the first row.}
    \vspace{-0.3cm}
    \resizebox{\linewidth}{!}{
    \begin{tabular}{c|c|c|c|c|c|c}
    \toprule
    ~ & \multirow{2}{*}{Attack} & Adv- & Adv- & Ens3- & Ens4- & Ens-  \\
    & & Inc-v3 & IncRes-v2 & Inc-v3 & Inc-v3 & IncRes-v2 \\
    \midrule    
    \multirow{15}{*}{\rot{Inc-v3}} & MIM & 23.2 & 17.6 & 14.6 & 16.1 & 7.5 \\
    ~ & DIM & 27.9 & 28.1 & 22.3 & 23.1 & 11.6 \\
    ~ & TIM & 31.9 & 26.3 & 30.9 & 31.2 & 22.2 \\
    ~ & PIM & 27.1 & 20.4 & 14.3 & 16.8 & 7.9 \\
    ~ & FIA & 57.3 & 57.8 & 45.2 & 46.5 & 26.2 \\
    ~ & VIM & 37.3 & 39.0 & 33.7 & 33.9 & 18.8 \\
    ~ & \bf FMAA & \bf 68.1 & \bf 69.3 & \bf 51.4 & \bf 49.8 & \bf 27.9 \\
    \cmidrule{2-7}
    ~ & PIDIM & 33.4 & 27.6 & 19.5 & 21.2 & 10.2 \\
    ~ & TIDIM & 46.7 & 40.9 & 44.4 & 45.7 & 33.5 \\
    ~ & PITIDIM & 50.8 & 43.0 & 46.1 & 49.8 & 34.8 \\
    ~ & FIA+PIDIM & 61.1 & 60.9 & 48.1 & 48.3 & 30.4 \\
    ~ & FIA+PITIDIM & 63.7 & 57.3 & 61.9 & 63.0 & 51.8 \\
    ~ & \bf FMAA+PIM & 69.4 & 71.1 & 48.2 & 49.3 & 24.3 \\
    ~ & \bf FMAA+PIDIM & 72.6 & \textbf{73.4} & 53.2 & 54.1 & 31.2 \\
    ~ & \bf FMAA+PITIM & 75.7 & 69.0 & 74.6 & \textbf{75.3} & 62.8 \\
    ~ & \bf FMAA+PITIDIM & \textbf{77.8} & 70.6 & \textbf{76.4} & 75.0 & \textbf{62.9} \\ \midrule    
    
    \multirow{15}{*}{\rot{IncRes-v2}} & MIM & 26.4 & 30.7 & 21.0 & 24.4 & 11.9 \\
    ~ & DIM & 33.9 & 43.3 & 32.7 & 31.1 & 20.5 \\
    ~ & TIM & 40.2 & 44.7 & 41.5 & 41.5 & 38.2 \\
    ~ & PIM & 30.7 & 29.0 & 19.0 & 20.4 & 10.9 \\
    ~ & FIA & 53.6 & 57.4 & 47.4 & 45.9 & 35.5 \\
    ~ & VIM & 45.0 & 60.5 & 48.6 & 46.0 & 38.2 \\
    ~ & \bf FMAA & \bf 74.5 & \bf 79.6 & \bf 67.2 & \bf 61.5 & \bf 51.4 \\
    \cmidrule{2-7}
    ~ & PIDIM & 35.7 & 42.8 & 28.0 & 27.7 & 16.0 \\
    ~ & TIDIM & 51.1 & 54.5 & 52.1 & 51.3 & 49.8 \\
    ~ & PITIDIM & 59.9 & 63.7 & 57.7 & 59.5 & 55.0 \\
    ~ & FIA+PIDIM & 59.6 & 62.5 & 51.0 & 47.2 & 37.3 \\
    ~ & FIA+PITIDIM & 56.1 & 53.0 & 54.7 & 56.7 & 51.1 \\
    ~ & \bf FMAA+PIM & 77.8 & \textbf{84.2} & 67.0 & 61.4 & 47.4 \\
    ~ & \bf FMAA+PIDIM & \textbf{79.8} & 82.7 & 68.1 & 62.2 & 47.1 \\
    ~ & \bf FMAA+PITIM & 77.5 & 78.4 & \textbf{78.8} & 76.4 & \textbf{75.0} \\ 
    ~ & \bf FMAA+PITIDIM & 77.6 & 78.6 & 78.1 & \textbf{77.1} & 74.6 \\ 
    \midrule    
    
    \multirow{15}{*}{\rot{Res-v1-152}} & MIM & 46.7 & 43.7 & 45.9 & 47.0 & 36.0 \\
    ~ & DIM & 67.3 & 63.6 & 65.0 & 65.5 & 54.5 \\
    ~ & TIM & 42.7 & 38.4 & 44.3 & 48.9 & 38.7 \\
    ~ & PIM & 52.1 & 48.8 & 51.9 & 50.0 & 38.1 \\
    ~ & FIA & 80.9 & 75.5 & 76.1 & 75.5 & 67.0 \\
    ~ & VIM & 72.5 & 69.6 & 71.1 & 71.4 & 62.5 \\
    ~ & \bf FMAA & \bf 85.7 & \bf 80.8 & \bf 80.4 & \bf 80.0 & \bf 70.1 \\
    \cmidrule{2-7}
    ~ & PIDIM & 72.3 & 69.3 & 68.7 & 67.8 & 56.3 \\
    ~ & TIDIM & 53.4 & 47.2 & 57.2 & 59.7 & 47.3 \\
    ~ & PITIDIM & 62.3 & 56.7 & 66.0 & 68.2 & 58.5 \\
    ~ & FIA+PIDIM & 88.2 & 86.3 & 84.6 & 83.7 & 77.0 \\
    ~ & FIA+PITIDIM & 68.8 & 65.8 & 70.4 & 74.7 & 65.5 \\
    ~ & \bf FMAA+PIM & 89.2 & 85.8 & 84.8 & 82.9 & 73.3 \\
    ~ & \bf FMAA+PIDIM & \textbf{91.7} & \textbf{90.1} & \textbf{88.8} & \textbf{87.8} & \textbf{81.8} \\
    ~ & \bf FMAA+PITIM & 71.7 & 68.3 & 73.1 & 76.5 & 66.5 \\ 
    ~ & \bf FMAA+PITIDIM & 72.3 & 69.1 & 75.4 & 78.7 & 67.4 \\ 
    \midrule    
    
    \multirow{15}{*}{\rot{Vgg-16}} & MIM & 75.7 & 68.8 & 76.3 & 73.2 & 60.8 \\
    ~ & DIM & 81.9 & 76.9 & 81.5 & 80.7 & 70.2 \\
    ~ & TIM & 55.1 & 46.3 & 55.9 & 58.8 & 44.8 \\
    ~ & PIM & 81.7 & 75.1 & 79.1 & 76.8 & 66.9 \\
    ~ & FIA & 92.8 & 90.6 & 91.7 & 91.7 & 86.3 \\
    ~ & VIM & 83.2 & 77.0 & 80.6 & 81.5 & 71.5 \\
    ~ & \bf FMAA & \bf 95.3 & \bf 91.6 & \bf 94.3 & \bf 94.2 & \bf 88.6 \\
    \cmidrule{2-7}
    ~ & PIDIM & 83.8 & 78.4 & 82.2 & 81.3 & 71.1 \\
    ~ & TIDIM & 60.9 & 52.6 & 62.4 & 65.8 & 52.1 \\
    ~ & PITIDIM & 66.2 & 57.6 & 66.6 & 70.0 & 56.8 \\
    ~ & FIA+PIDIM & 95.6 & 93.8 & 94.7 & 94.3 & 90.6 \\
    ~ & FIA+PITIDIM & 76.0 & 72.6 & 79.2 & 82.5 & 72.4 \\
    ~ & \bf FMAA+PIM & 96.7 & 93.6 & 95.8 & 94.9 & 90.4 \\
    ~ & \bf FMAA+PIDIM & \textbf{97.0} & \textbf{94.5} & \textbf{96.5} & \textbf{96.2} & \textbf{91.0} \\
    ~ & \bf FMAA+PITIM & 74.4 & 68.8 & 78.2 & 79.7 & 68.6 \\ 
    ~ & \bf FMAA+PITIDIM & 75.2 & 71.6 & 79.5 & 81.9 & 70.8 \\ 
    \bottomrule
    \end{tabular}}
    \label{tab:attack_defense}
    \vspace{-0.4em}
\end{table}

\smallskip
\noindent{\bf Attacking Ensemble Models.} Moreover, we study the effectiveness of our method using an ensemble of models following the settings in work \cite{liu2016delving}. Specifically, we ensemble Vgg-16, Vgg-19, Res-v1-50 and Res-v1-152 as source models to attack defense models. The results are shown in Table \ref{tab:attack_ensemble}, which reveals that our method improves the performance by $3.6\%$.

\renewcommand\arraystretch{0.9}
\begin{table}[!ht]
    \centering
    \small
    \caption{\small Success rate ($\%$) of different methods using ensemble model (Res-v1-50, Res-v1-152, Vgg-16, and Vgg-19) to attack different defense models.}
    \vspace{-0.3cm}
    \resizebox{0.9\linewidth}{!}{
    \begin{tabular}{c|c|c|c|c|c}
    \toprule
    \multirow{2}{*}{Attack} & Adv- & Adv- & Ens3- & Ens4- & Ens-  \\
    & Inc-v3 & IncRes-v2 & Inc-v3 & Inc-v3 & IncRes-v2 \\
    \midrule
    MIM & 84.9 & 81.0 & 84.3 & 84.3 & 75.0 \\
    DIM & 94.2 & 91.8 & 93.4 & 92.9 & 89.0 \\
    TIM & 73.7 & 67.8 & 74.2 & 76.3 & 66.2 \\
    PIM & 88.8 & 83.8 & 86.8 & 86.2 & 76.7 \\
    TIDIM & 80.5 & 74.2 & 80.5 & 82.8 & 74.5 \\
    PITIDIM & 85.6 & 81.6 & 86.2 & 89.2 & 80.3 \\
    FIA & 93.5 & 89.8 & 90.4 & 90.4 & 82.5 \\
    VIM & 93.7 & 90.1 & 91.8 & 92.9 & 87.0 \\
    \bf FMAA & \textbf{97.1} & \textbf{96.3} & \textbf{96.9} & \textbf{96.0} & \textbf{93.2} \\ 
    \bottomrule
    \end{tabular}}
    \label{tab:attack_ensemble}
    \vspace{-0.4em}
\end{table}

\subsection{Ablation Study}
This section studies different settings of our method, including different feature-momentum weights, different layers to attack, different drop probabilities. 

\begin{figure}[ht!]
\centering
\vspace{-0.5em}
\includegraphics[width=\linewidth]{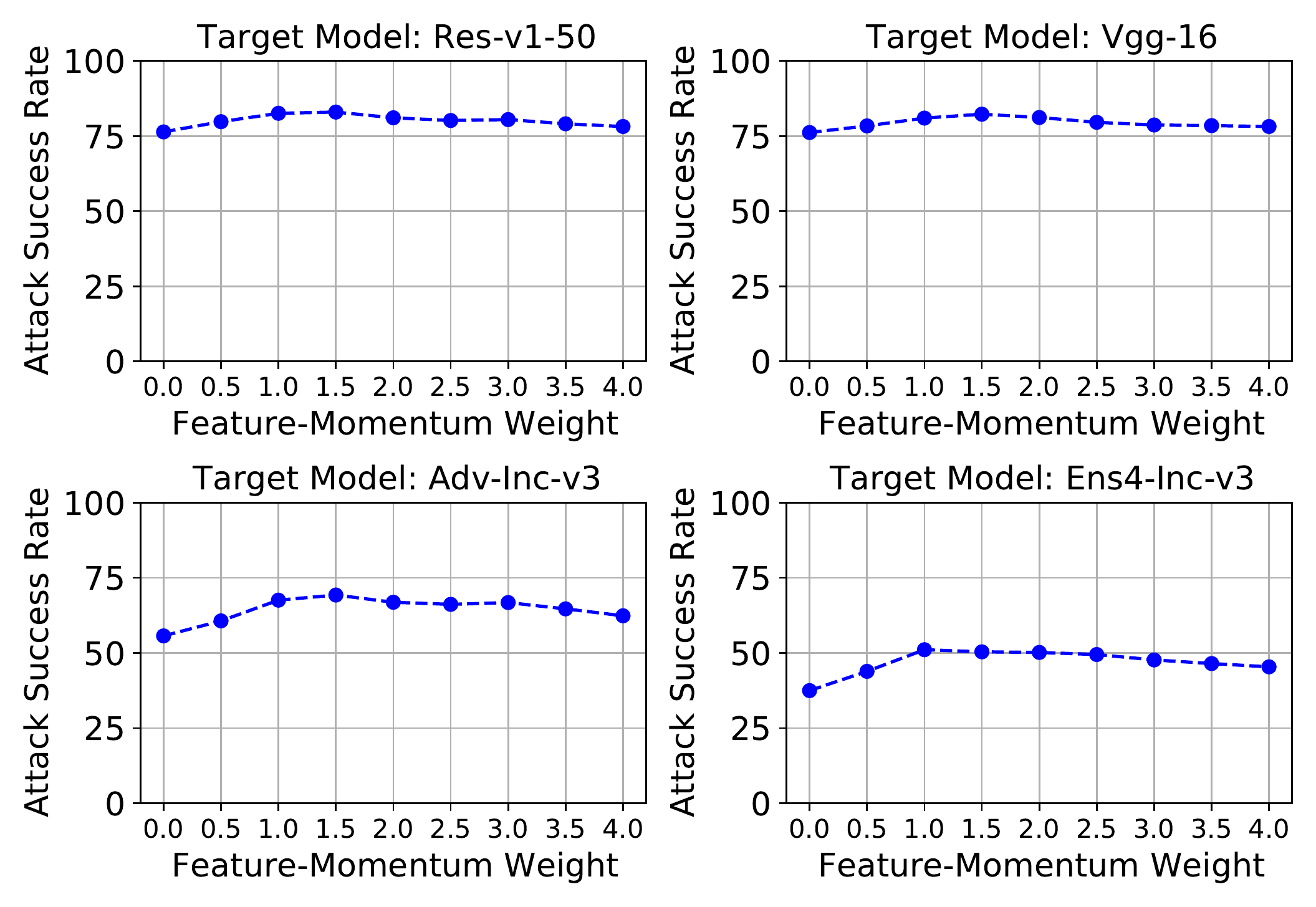}
\vspace{-0.7cm}
\caption{\small  The effect of feature-momentum weight $\lambda$ on attack success rate. The top figures show the performance of attacking two original models Res-v1-50 and Vgg-16, while the bottom ones correspond to defense models Adv-Inc-v3 and Ens4-Inc-v3.}
\label{fig:fm_weight}
\vspace{-0.3cm}
\end{figure}

\smallskip
\noindent{\bf Effect of Feature-Momentum Weight.}
This part investigates the effect of different feature-momentum weight $\lambda$. The source model of this experiment is Inc-v3 and the $\lambda$ is changed from $0$ to $4.0$ with the step of $0.5$. Fig. \ref{fig:fm_weight} illustrates the effect of using different $\lambda$ to attack IncRes-v2, Vgg-16, Adv-Inc-V3, and Ens4-Inc-v3. Note the weight $\lambda = 0$ denotes no momentum is applied. We can observe that the curve on different target models shows a similar trend that the attack success rate raises with weight increasing and then drops when the weight turns too large. For all target models, the range $[1.0, 1.5]$ is an optimal choice.

\begin{figure}[ht!]
\centering
\vspace{-0.3cm}
\includegraphics[width=0.9\linewidth]{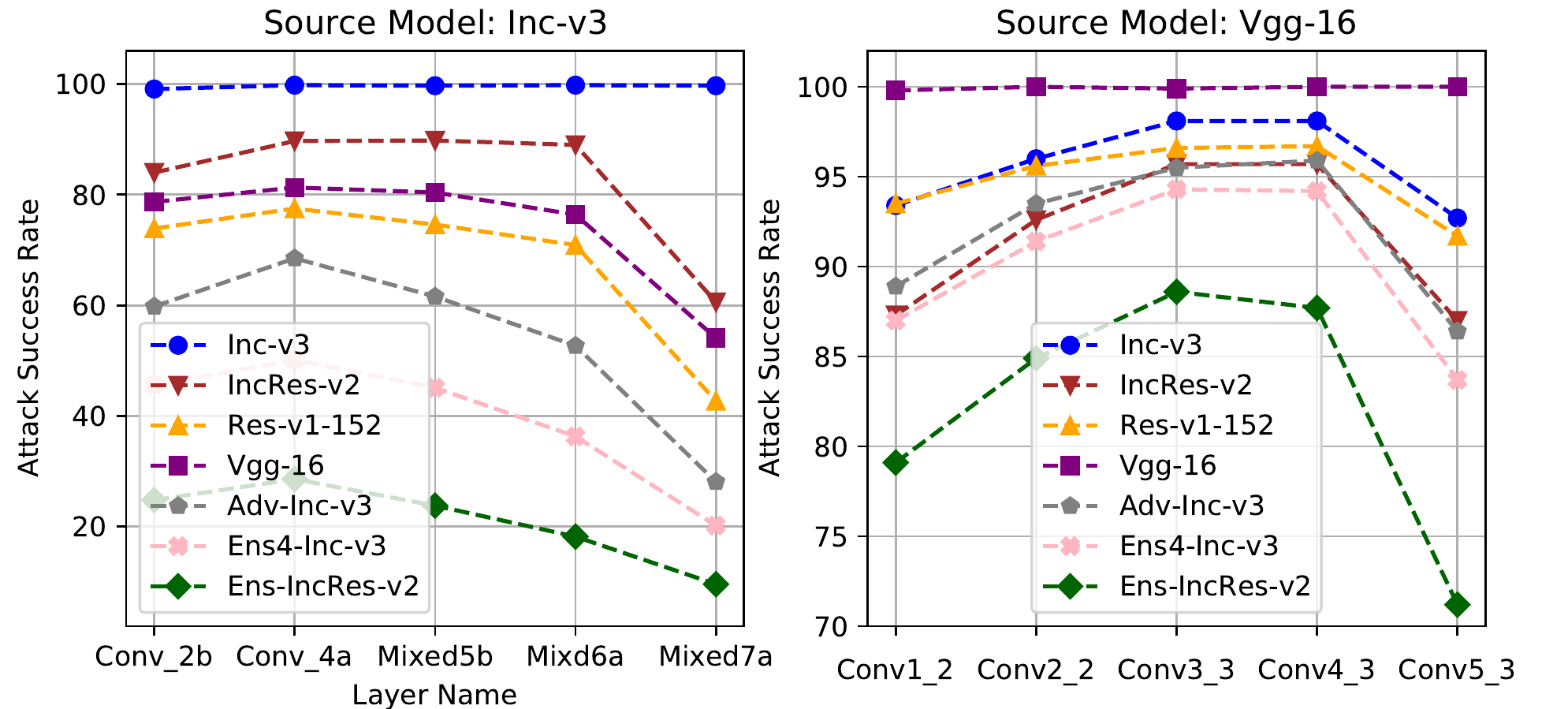} \\
\hspace{-0.3cm}
\includegraphics[width=0.85\linewidth]{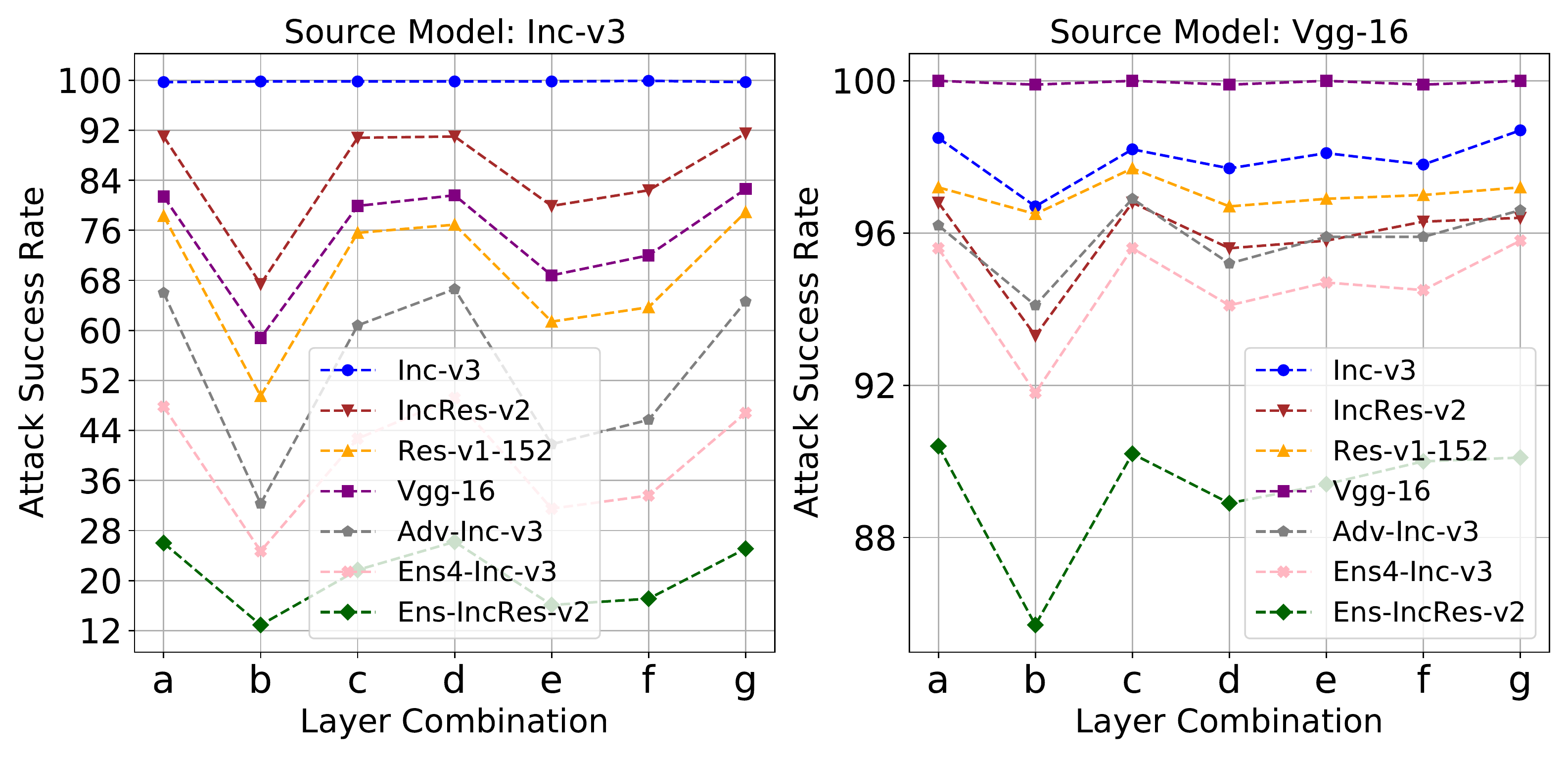}
\vspace{-0.3cm}
\caption{\small  The Effect of various layers on attack success rate. The top figures show the effect of using different single layers, while the bottom ones show the effect of using different layer combinations. }
\label{fig:layers}
\vspace{-0.3cm}
\end{figure}

\smallskip
\noindent{\bf Attacking Various Layers.} We study the effect of attacking various layers in this part. Specifically, we choose Inc-v3 and Vgg-16 as the source models. As the shallow and deep layers contain different levels of features, we select the layers from shallow to deep and also define several layer combinations for comprehensive study. For source model Inc-v3, we select layers of Conv\_2b, Conv\_4a, Mixed5b, Mixed6a and Mixed7a. For source model Vgg-16, we select layers of Conv1\_2, Conv2\_2, Conv3\_3, Conv4\_3 and Conv5\_3. For layer combinations, we select seven combinations as follows: (Conv\_4a, Mixed5b), (Conv\_2b, Mixed7a), (Conv\_4a, Mixed6a), (Conv\_2b, Mixed5b), (Mixed5b, Mixed7a), (Conv\_2b, Mixed5b , Mixed7a), (Conv\_4a, Mixed5b, Mixed6a) for Inc-v3 and (Conv3\_3, Conv4\_3), (Conv1\_2, Conv5\_3), (Conv2\_2, Conv4\_3), (Conv1\_2, Conv3\_3), (Conv3\_3, Conv5\_3), (Conv1\_2, Conv3\_3, Conv5\_3), (Conv2\_2, Conv3\_3, Conv4\_3) for Vgg-16. These combinations are denoted as layer $a \sim g$. Fig. \ref{fig:layers} (top) shows the performance of attacking various layers, which reveals that attacking the middle layer can achieve the best performance for both of source models. Fig. \ref{fig:layers} (bottom) shows the performance of attacking various layer combinations. We can observe that the combination \emph{a} and \emph{g}, which includes the layers in the middle of the network, perform best.

\begin{figure}[ht!]
\centering
\vspace{-0.3cm}
\includegraphics[width=\linewidth]{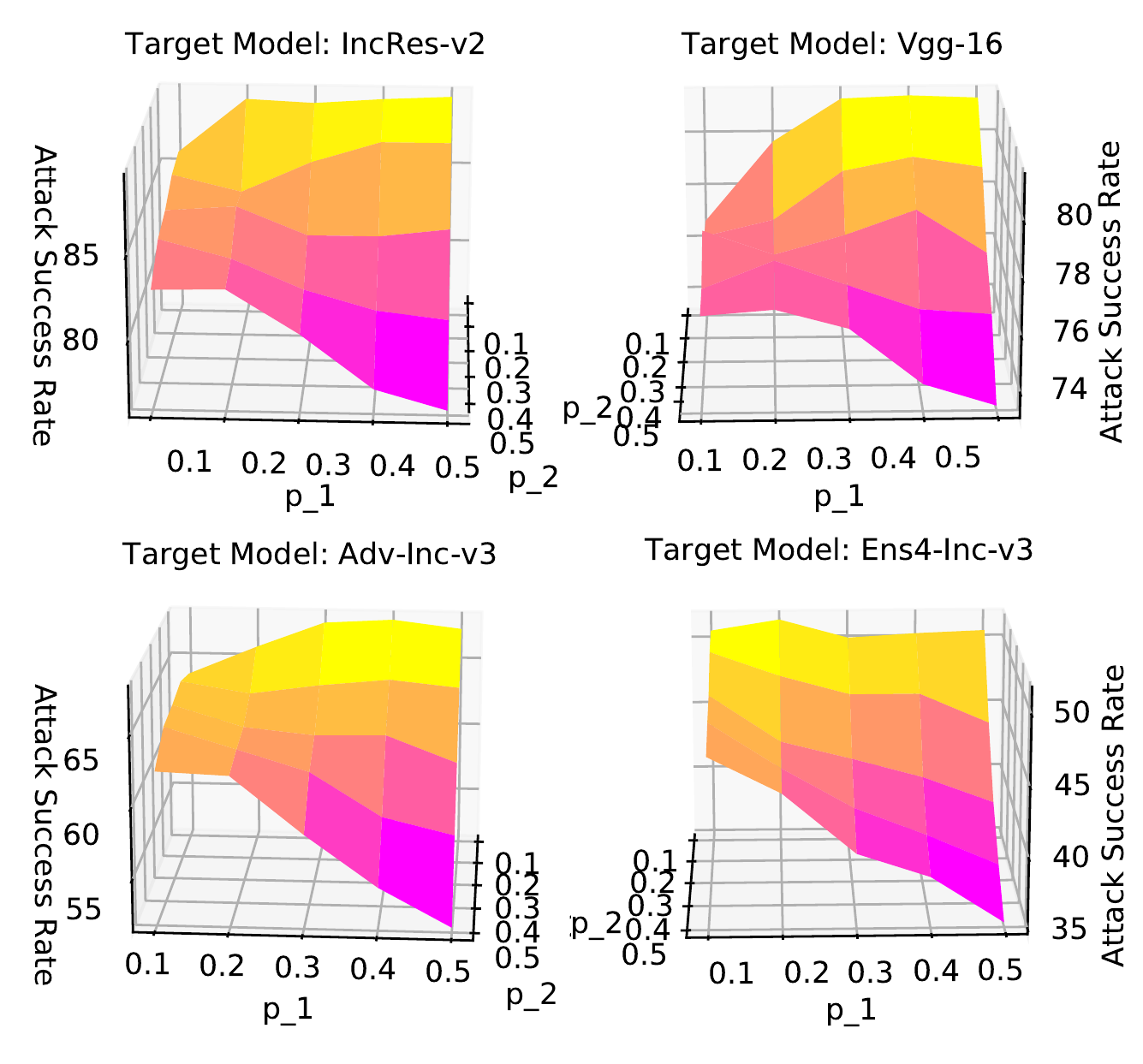}
\vspace{-0.8cm}
\caption{\small Effect of the drop probabilities on attack success rate.}
\label{fig:drop}
\vspace{-0.5cm}
\end{figure}

\smallskip
\noindent{\bf Effect of Drop Probability.} This part studies the effect of different drop probability in random pixel masking. As aforementioned, we use two drop probabilities at the start iteration and the rest iterations. We denote these two probabilities as $p_1$ and $p_2$.
We change these two probabilities from $0.1$ to $0.5$ with the step of $0.1$. As shown Fig. \ref{fig:drop}, the top and bottom denote the performance of attacking two original models (IncRes-v2 and Vgg-16), and two defense models (Adv-Inc-v3 and Ens4-Inc-v3). We can observe that a large $p_1$ with a small $p_2$ can perform well, which is probably because the random transformation at the beginning can affect the following iterations, such that a small $p_2$ is sufficient.


\section{Conclusion}
This paper describes a new method called feature-momentum adversarial attack to improve transferability. Specifically, our method estimates a dynamic guidance map for feature disturbance using momentum, which considers the knowledge from previous iterations, leading to a notable enhancement on transferability. Our method is evaluated on several original and defense models with comparison to several state-of-the-art methods, which demonstrates the superiority of our method.

{\small 
\bibliographystyle{named}
\bibliography{ref}
}
\end{document}